%% file: hd3-2015_v3.tex
\documentclass[a4paper,amsmath,amssymb]{jpconf}
\usepackage{graphicx}
\usepackage{amsmath, amsthm, amssymb, dsfont}

\input{commands}

\begin{document}

\title{Performance of a community detection algorithm based on semidefinite programming}

\author{Federico Ricci-Tersenghi}

\address{Dipartimento di Fisica, INFN--Sezione di Roma1 and CNR--Nanotec, Universit\`a La Sapienza, Piazzale Aldo Moro 5, I-00185 Roma, Italy.}

\ead{federico.ricci@uniroma1.it}

\author{Adel Javanmard}

\address{USC Marshall School of Business, University of Southern California}

\author{Andrea Montanari}

\address{Department of Electrical Engineering and Department of Statistics, Stanford University}

\begin{abstract}
The problem of detecting communities in a graph is maybe one the most studied inference problems, given its simplicity and widespread diffusion among several disciplines.
A very common benchmark for this problem is the stochastic block model or planted partition problem, where a phase transition takes place in the detection of the planted partition by changing the signal-to-noise ratio.
Optimal algorithms for the detection exist which are based on spectral methods, but we show these are extremely sensible to slight modification in the generative model.
Recently Javanmard, Montanari and Ricci-Tersenghi~\cite{sdp} have used statistical physics arguments, and numerical simulations to show that finding communities in the stochastic block model via semidefinite programming is quasi optimal. Further, the resulting semidefinite relaxation can be solved efficiently, and is very robust with respect to changes in the generative model.
In this paper we study in detail several practical aspects of this new algorithm based on semidefinite programming for the detection of the planted partition. The algorithm turns out to be very fast, allowing the solution of problems with $O(10^5)$ variables in few second on a laptop computer.
\end{abstract}

\section{Introduction and model definition}
\vspace{3mm}

When dealing with a high-dimensional dataset one often looks for hidden structures, that may be representative of the signal one is trying to extract from the noisy dataset.
In the community detection problem, we are asked to find the most significative clustering of the vertices of a graph, such that intra-cluster connections are much more abundant/sparse in the assortative/disassortative case with respect to inter-clusters connections (see Ref.~\cite{fortunato2010community} for a review).
A common benchmark in this class is provided by the the so-called planted partition problem or stochastic block model (SBM) \cite{holland1983stochastic}, defined as follows: an undirected graph $G=(V,E)$ is given, where $V=\{1,\ldots,n\}$ is the vertex set and $E$ is the edge set. Vertices are divided in $q$ equal size groups $V_k$, with $k=1,\ldots,q$, such that $V=\cup_{k=1}^q V_k$, $|V_k|=n/q$ and $V_i \cap V_j = \emptyset$ if $i\neq j$. The function $\kappa(i)$ returns the cluster vertex $i$ belongs to. Conditioned on the partition $\{V_k\}_{k=1,\ldots,q}$, edges are drawn independent with
\[
\prob\Big[(i,j)\in E\Big|\{V_k\}\Big] = \left\{
\begin{array}{ll}
\cin/n \quad & \text{ if } \kappa(i) = \kappa(j)\,,\\
\cout/n \quad & \text{ if } \kappa(i) \neq \kappa(j)\,.
\end{array}
\right.
\]
The resulting random graph is sparse and has a mean degree equal to $c = [\cin + (q-1) \cout]/q$.
We will be mainly concerned with the assortative case, where $\cin>\cout$ holds strictly. 
The goal is to detect the partition $\{V_k\}$ given the graph $G$.

The Bayesian approach \cite{decelle2011asymptotic} predicts the existence of a threshold at
\[
\cin - \cout = q \sqrt{c}\,,
\]
for $q\le 4$, separating a phase where detecting the partition is impossible from a phase where the detection can be achieved, using the best possible algorithms.
We defined the signal-to-noise ratio as
\[
\lambda = \frac{\cin - \cout}{q \sqrt{c}}\,,
\]
such that the phase transition for $q\le 4$ takes place at $\lambda_c=1$.
We will mainly be interested in the $q=2$ case, where the existence of the threshold at $\lambda_c=1$ has been proved rigorously \cite{mossel2013proof,massoulie2014community}.
In this case the planted partition can be conveniently coded in a vector $\bx_0 \in \{+1,-1\}^n$, and, calling $\hbx(G)$ the estimate of the partition in graph $G$ obtained by any inference procedure, we can quantify the success of the detection algorithm computing the overlap with respect to the planted partition (i.e.\ the absolute value of the normalized scalar product)
\[
Q = \frac1n |\<\hbx(G), \bx_0\>|\,.
\]
Eventually we can consider also its average value over the ensemble of random graphs, $\E[Q]$.
For the SBM Ref.~\cite{decelle2011inference} proposed a belief-propagation message-passing algorithm to find the Bayes optimal estimator, which has indeed a non-zero overlap with the planted partition as soon as $\lambda>\lambda_c=1$.

Spectral methods based on the Laplacian (unnormalized or normalized) are known to be sub-optimal, since they have a threshold which is strictly larger than the optimal one, even for regular graphs, namely \ $\lambda_c^\text{Lap}=\sqrt{c/(c-1)}>1$ \cite{kawamoto2015limitations}.
However, a new spectral method based on the non-backtracking matrix introduced in Ref.~\cite{krzakala2013spectral} achieves optimality in the detection of the planted partition in the SBM, at the cost of computing the complex spectrum of a non symmetric matrix.
Later, such a spectral method has been strongly simplified by showing its similarity to the computation of the spectrum of the so-called Bethe Hessian matrix \cite{saade2014spectral}, which is a $n\times n$ symmetric matrix defined as
\[
H(r) = (r^2-1)\mathds{1} - rA - D\,,
\]
where $A$ is the adjacency matrix, $A_{ij}=A_{ji}=\mathbb{I}[(ij)\in E]$, and $D$ is a diagonal matrix with entries equal to the vertex degrees $d_i$. In the assortative SBM with $q=2$, the planted partition is detected by computing the negative eigenvalues of $H(\sqrt{c})$ \cite{saade2014spectral} and the best estimator $\hbx^\text{\tiny BH}(G)$ turns out to be given by the vector of signs of the components of the eigenvector corresponding to the second largest (in absolute value) eigenvalue.

\section{Spectral methods versus optimization methods}
\vspace{3mm}

\begin{figure}[ht]
\begin{center}
\includegraphics[width=0.9\textwidth]{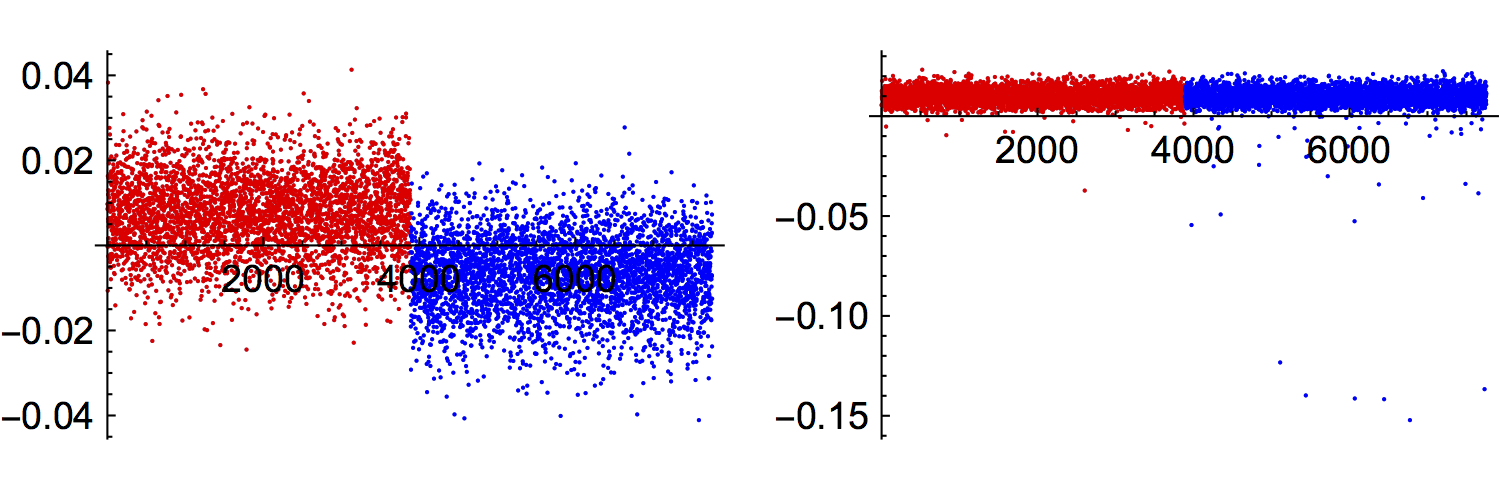}
\end{center}
\caption{\footnotesize Components of the Bethe Hessian estimator for the planted partition in a graph of $7\,755$ vertices and $13\,348$ edges generated according to the SBM and then reduced to its 2-core (left). In the right plot we have used the same graph, where 2 cliques of sizes 3 and 5 has been added. A difference of only 13 edges not generated according to the SBM makes the spectral method completely useless for the detection.
}
\label{fig:BHfailure}
\end{figure}

Although the partition detection based on the Bethe Hessian is optimal for the SBM, it turns out to be not very robust if the generative model departs even slightly from the random graph ensemble defined above.
Given that the underlying assumption in the SBM is that the graph is locally tree-like, the addition of few cliques may drastically deteriorate the performance of the algorithm based on the Bethe Hessian.
Just to give you an idea of how fragile may be the Bethe Hessian based estimator $\hbx^\text{\tiny BH}(G)$ we show in Fig.~\ref{fig:BHfailure} its components for 2 almost identical graphs. The graph in the left plot is a typical graph from the SBM ensemble with $q=2$, $c=3$, $\lambda=1.2$ and $n=10^4$ (actually we focus on the 2-core of the graph, for reasons explained below, that has $7\,755$ vertices and $13\,348$ edges in this case): the resulting overlap with the planted partition is $Q=0.59$, in agreement with previous studies \cite{saade2014spectral}. The graph in the right plot is exactly the same graph in the left plot with the addition of \emph{only two cliques} of sizes 3 and 5 (that is just 13 more edges!), but the ability to infer the planted partition from the spectrum of the Bethe Hessian is completely lost, as a consequence of the eigenvector localization. Indeed the resulting overlap is $Q=0.01$.

Robustness is an important requirement, since no generative model is exact in applications.
To this aim, converting the inference problem to an optimization problem is welcome.
Obviously the result will depend on the objective function to maximize.
One of the most used objective functions for detecting communities in real world graphs is the modularity \cite{clauset2004finding,newman2006modularity} defined as
\[
\sum_{g = 1}^q \sum_{i,j \in g} \left( A_{ij} - \frac{ d_i d_j }{ 2m } \right)\,,
\]
where the index $g$ runs over the clusters of vertices in the partition. Modularity measures the excess of intra-cluster connections with respect to a random assignment. Maximizing the modularity is not an easy job, especially because one can find several phase transition in the space of partitions \cite{zhang2014scalable,schulke2015multiple}, which are likely to affect the the maximum likelihood problem.

In the case of $q=2$ groups of equal size, the problem of detecting the graph partition is equivalent to the minimum bisection problem, and the objective function to be minimized is just the cut size.
Consequently the maximum-likelihood estimator is given by
\beq
\xml(G) = \mathop{\argmax}_{\bx \in \{+1,-1\}^n} \Big( \sum_{(ij)\in E} x_i x_j \Big | \sum_i x_i = 0 \Big)\,.
\label{eq:ml}
\eeq
However computing the estimator in \eqref{eq:ml} is a hard problem (NP-complete and non polynomial time approximable in the worst case \cite{arora2005non}), because the function to be maximized may have several local maxima, that trap the optimization algorithms.

In order to approximate this problem with a more tractable version it is customary to relax it as semidefinite programming (SDP) over the set of $n \times n$ real and symmetric matrices $\bC$
(see, e.g.  \cite{goemans1995improved,lovasz1991cones} and, in the present context \cite{abbe2016exact,guedon2015community}):
\beq
\text{maximize}\;\sum_{i,j} A_{ij} C_{ij} \qquad \text{subject to}\;\; \bC \succeq 0\,,\;
C_{ii} = 1\,,\; \sum_j C_{ij} = 0 \;\forall i\,.
\label{eq:sdp}
\eeq
The positive semidefinite condition, $\bC \succeq 0$, requires all the eigenvalues of $\bC$ to be non-negative and makes the feasible set convex, thus ensuring the existence of a tractable maximizer of \eqref{eq:sdp}.

The maximum-likelihood problem \eqref{eq:ml} is recovered from the formulation in \eqref{eq:sdp} by enforcing the matrix $\bC$ to be of rank 1: $C_{ij}=x_i x_j$.
So, in general, willing to solve the non-convex problem in \eqref{eq:sdp} with rank 1 matrices, one may search for a solution to the convex problem with generic rank $n$ matrices, and then project back this solution to the space of rank 1 matrices.

A convenient way to represent a $n \times n$ real and symmetric positive semidefinite matrix of rank $m$ is to consider it as a correlation matrix between $n$ real vectors of $m$ components each:
\beq
C_{ij} = \ux_i \cdot \ux_j\,,\quad \text{with}\; \ux_i \in \reals^m\,,\; \|\ux_i\|^2 = \ux_i \cdot \ux_i = 1\,.
\label{eq:Cm}
\eeq

\section{Our community detection algorithm and it performances}
\vspace{3mm}

In order to solve problem \eqref{eq:sdp} over the set of rank $m$ matrices we have recently proposed the following procedure \cite{sdp}. First of all search for a configuration of the $n$ unit-length $m$-components vectors $\ux_i\in \reals^m$, $\|\ux_i\|=1$, optimizing the following objective function
\beq
\text{maximize}\;\; \sum_{(ij)\in E} \ux_i \cdot \ux_j\,, \qquad \text{subject to}\;\;\sum_i \ux_i = \underline{0}\,.
\label{eq:oursdp}
\eeq
Let us call $\ubx^*=\{\ux_1^*,\ldots,\ux_n^*\}$ the maximizer.
To project back the maximizer to a vector of $n$ reals, we first compute the matrix $\Sigma \in \reals^{m \times m}$ measuring the correlations among the $m$ components of the maximizer averaged over the entire graph
\beq
\Sigma_{jk} = \frac1n \sum_{i=1}^n (\ux_i^*)_j (\ux_i^*)_k\,,
\label{eq:Sigma}
\eeq
whose principal component we call $\uv_1$, and then we project each $\ux_i$ over $\uv_1$.
Thus the rank $m$ SDP estimator $\xsdp(G)$ has components
\beq
\hat{x}_i^{\mbox{\tiny{SDP}}}(G) = \sign(\ux_i \cdot \uv_1)\,.
\label{eq:proj}
\eeq
As before we measure the success of our detection algorithm by computing the overlap with respect to the planted partition
\[
\qsdp = \frac1n |\< \xsdp, \bx_0 \>|\,.
\]

In Ref.~\cite{sdp} we have shown that the above algorithm, in the $m\to\infty$ limit, is optimal for synchronization problems defined on dense graphs and almost optimal for solving the SBM. By `almost optimal' we mean that the overlap $\qsdp$ shows a phase transition at $\lsdp$ slightly larger than the optimal $\lambda_c=1$: for example for $c=3$ we have $\lsdp \simeq 1.017$, with $\qsdp >0$ for $\lambda>\lsdp$.
Just for comparison, we remind that for $c=3$ the spectral methods based on the Laplacian are unable to detect the planted partition as long as $\lambda < \sqrt{3/2} \simeq 1.22$.

The algorithm we use to find the maximizer in \eqref{eq:oursdp} is block-coordinate descent. In physics language a zero temperature dynamics for a model with $m$-component spins $\ux_i$ placed on the vertices of the graph $G$, with the addition of an external field that self adapt in order to keep the global magnetization null. In practice at each step of the algorithm we update spins in a random order aligning each spin to its local field
\[
\ux_i^{(t+1)} = \frac{\sum_{j:(ij)\in E} \ux_j^{(t)}-\underline{M}(t)}{\|\sum_{j:(ij)\in E} \ux_j^{(t)}-\underline{M}(t)\|}\,,
\]
with the global magnetization being $\underline{M}(t)=\sum_i \ux_i$. We always check that the global magnetization becomes very small at large times, $\lim_{t\to\infty}\underline{M}(t)=\underline{M}(\infty)\ll 1$.
As a stopping criterion we check the largest variation in a spin during the last step, $\Delta_\text{max} = \max_i \|\ux_i^{(t+1)} - \ux_i^{(t)}\|$, and we stop when $\Delta_\text{max}<\varepsilon$, with $\varepsilon=10^{-3}$ or $10^{-4}$ (the specific value is rather irrelevant, being the results independent of $\varepsilon$ as long as $\varepsilon \lesssim 10^{-3}$). We call $\tconv$ the number of steps required to meet the stopping criterion. Earlier literature on rank-constrained SDPs
uses different approaches \cite{burer2003nonlinear,journee2010low,boumal2014manopt}. We found that the block-coordinate descent method studied here is significantly faster. 

The main parameter in this algorithm is the number of components $m$, and we expect the behavior of the algorithm to depend strongly on $m$ at least close to the critical point $\lambda_c=1$. Indeed for $m=1$ the maximization of the likelihood is a NP-hard problem. We expect a greedy dynamics, as the one we are using, to get stuck in some local maxima, while for $m\to \infty$ the problem is convex and thus the greedy dynamics should be able to reach the maximum.

Given that each step of the dynamics requires $O(m\,n)$ operations, an interesting aspect to study is the minimal value of $m$ that allows the algorithm to get close to the $m=\infty$ maximizer, such that the quasi optimal solution of Ref.~\cite{sdp} can then be obtained through the projection in \eqref{eq:proj}.
On general grounds, since the SDP in \eqref{eq:sdp} has $n$ constraints, one can show \cite{burer2003nonlinear} that for $m\ge 2\sqrt{n}$ the objective function in \eqref{eq:oursdp} has no local maxima, which are not global maxima
(under suitable conditions on the undrrlying graph). However, we expect $\sqrt{n}$ to be a loose upper bound for the optimal value of $m$ on random instances. Indeed, for $\lambda=0$ the model we are studying is close to an $m$-components spin glass on a random graph, whose ground state (i.e.\ the maximizer) has a number of non-zero components (i.e.\ the rank) growing roughly as $n^\mu$, with $1/3\lesssim\mu<2/5$ depending on the mean degree $c$ \cite{braun2006m}.
From the recent work \cite{montanari2015semidefinite} we known that the maximum of the objective function on rank $m$ solutions deviates for the corresponding $m=\infty$ value by no more than $O(1/m)$ terms. We will see that actually the convergence on the SDP estimator $\xsdp$ is much faster.

\begin{figure}[ht]
\begin{center}
\includegraphics[width=0.6\textwidth]{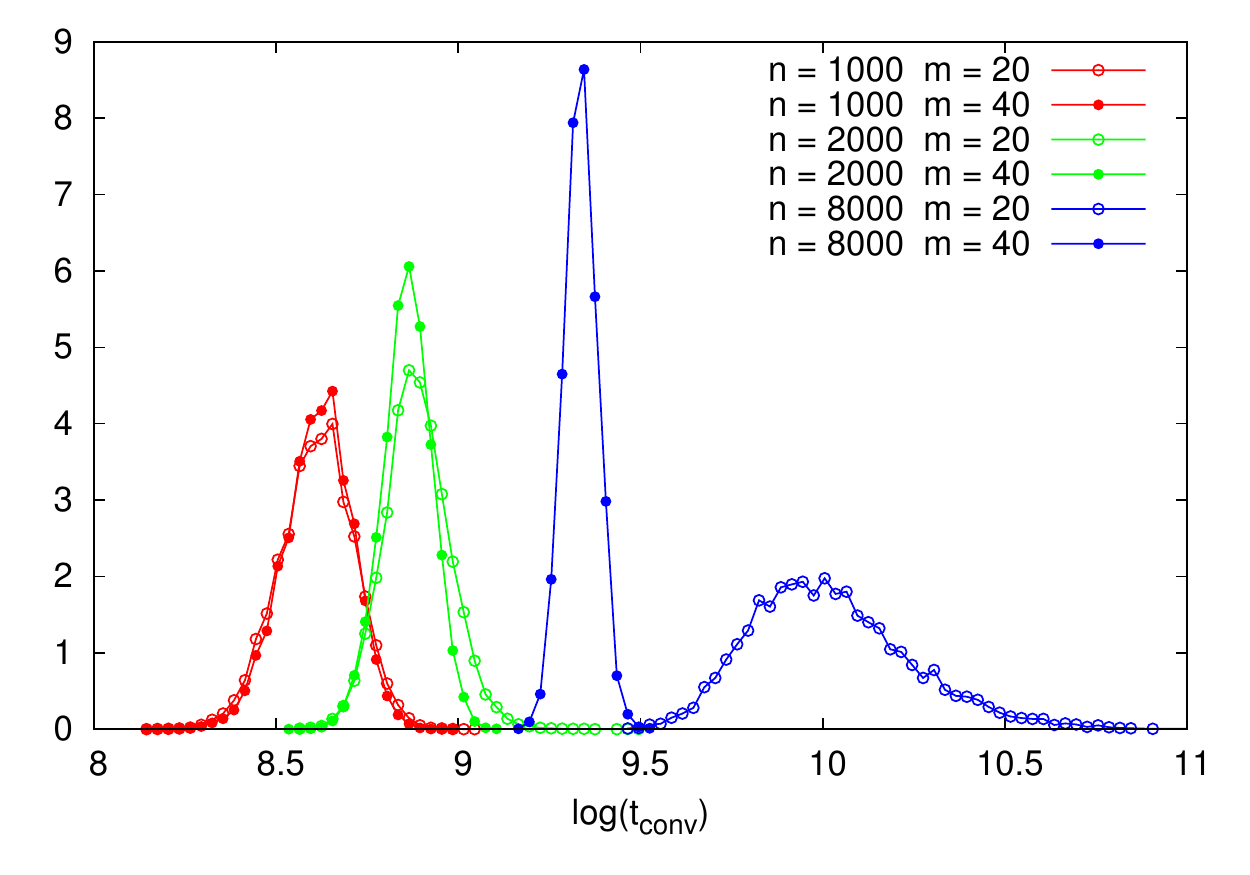}
\end{center}
\caption{\footnotesize Distribution of convergence times for our greedy algorithm measured on many graphs generated according to the SBM with $c=5$ and $\lambda=1$. The typical running time grows mildly with the problem size $n$, but if $m$ is not large enough the convergence time becomes much larger and sample to sample fluctuations much more severe.
}
\label{fig:convTimes}
\end{figure}

In Ref.~\cite{sdp} we have studied the effect of changing $m$ in solving the SBM close to the critical value $\lambda_c=1$. The kind of results we got are illustrated in Fig.~\ref{fig:convTimes} for graphs generated according to the SBM with $c=5$ and $\lambda=1$. We plot the histogram of the convergence times, $\tconv$, and we see that running the algorithm with $m=20$ and $m=40$ leads essentially to the same dynamics and the same convergence times for $n=1000$ and $n=2000$. On the contrary for $n=8000$ convergence times with $m=20$ are sensibly larger and much more disperse than with $m=40$ (notice we are plotting $\log(\tconv)$). Our conclusion is that the objective function we are climbing with the greedy dynamics is less smooth for $m=20$ than for $m=40$.

Although the dynamics get slower (i.e.\ $\tconv$ increases) reducing the value of $m$, because the objective function to be maximized gets rougher, each single step of the algorithm becomes much more economic, and so it is worth asking how much the maximizer reached by running the algorithm with a ``too small'' value of $m$ is informative for the goal of detecting communities.

The graphs that we use are generated according to the SBM with $c=3$, and subsequently they are pruned by eliminating recursively dangling ends. Equivalently we reduce the graph to its 2-core, and this has the advantage of reducing the noise level, without changing the complexity of the problem. Indeed any removed tree joins the core on a single vertex and the optimal solution is to assign all tree vertices to the same cluster the core vertex belongs to.

In order to study the robustness of our approach, we add cliques according to the following rule: for each vertex, with probability $p$, we add a clique among all its neighbour vertices. This superimposed structure is well justified since in many real world networks, especially social networks, structures of this kind are very common (two friends of a person tend to be also friends). But from the point of view of the SBM this is an additional perturbation which may degrade the performance of detecting algorithms. Indeed this is what we have shown preliminary in Fig.~\ref{fig:BHfailure} where the graph of the right plot has been generated with a very small $p=10^{-4}$ and still the effect on the Bethe Hessian algorithm is dramatic.

In order to get statistics on a single graph, we run the algorithm with 100 different \emph{clones}. Each clone starts from a different random initial condition and evolve according to the greedy algorithm illustrated above. Remind that the algorithm, although greedy, is not deterministic, because variables are updated in a random order. At all the times when we take measurements we compute the SDP estimator $\xsdp$ and the corresponding overlap $\qsdp$ for each of the 100 clones.

\begin{figure}[ht]
\begin{center}
\includegraphics[width=0.7\textwidth]{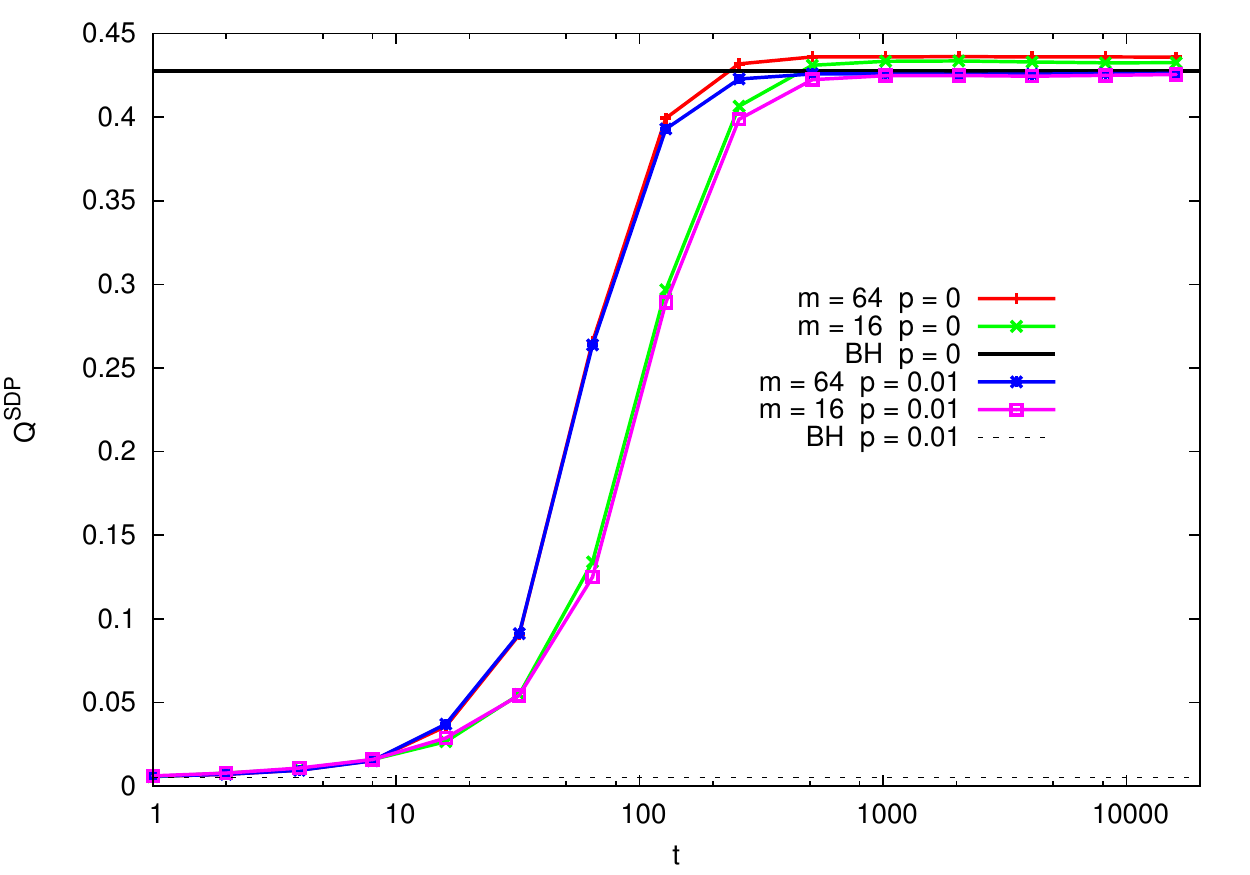}
\end{center}
\caption{\footnotesize Overlap with the planted partition obtained from the configurations at time $t$ in our algorithm. We report the average over the 100 clones. The comparison with the overlap achieved by the Bethe Hessian algorithm is favorable, especially when a small fraction of cliques is added, and the prediction from the Bethe Hessian is equivalent to a random guess.
The algorithm run with $m=64$ and $m=16$ requires different running times, but achieves essentially the same inference accuracy.
}
\label{fig:0}
\end{figure}

In Fig.~\ref{fig:0} we show the robustness of our algorithm with respect to the addition of cliques. We consider a graph generated initially with $n=40\,000$ nodes at $\lambda=1.1$: the resulting core has $30\,597$ vertices, $53\,161$ edges for $p=0$ and $54\,734$ edges for $p=0.01$.
In Fig.~\ref{fig:0} we plot $\qsdp$ as a function of the number of iteration of our algorithm, for $m=16$ and $m=64$. The horizontal lines report the value of the overlap $q$ achieved by the Bethe Hessian algorithm, which is very close to the Bayes optimal value, according to Ref.~\cite{saade2014spectral}.
We observe that for $p=0$ our algorithm reaches a higher value of the overlap with respect to the Bethe Hessian result, both for $m=16$ and $m=64$. More importantly the addition of cliques ($p=0.01$ case) deteriorates very little the performances of our algorithm, while the spectral method return a random guess with $Q \simeq 0$.
By virtue of robustness of our algorithm, hereafter we are going to study only the $p=0$ case.

\begin{figure}[ht]
\begin{center}
\includegraphics[width=0.7\textwidth]{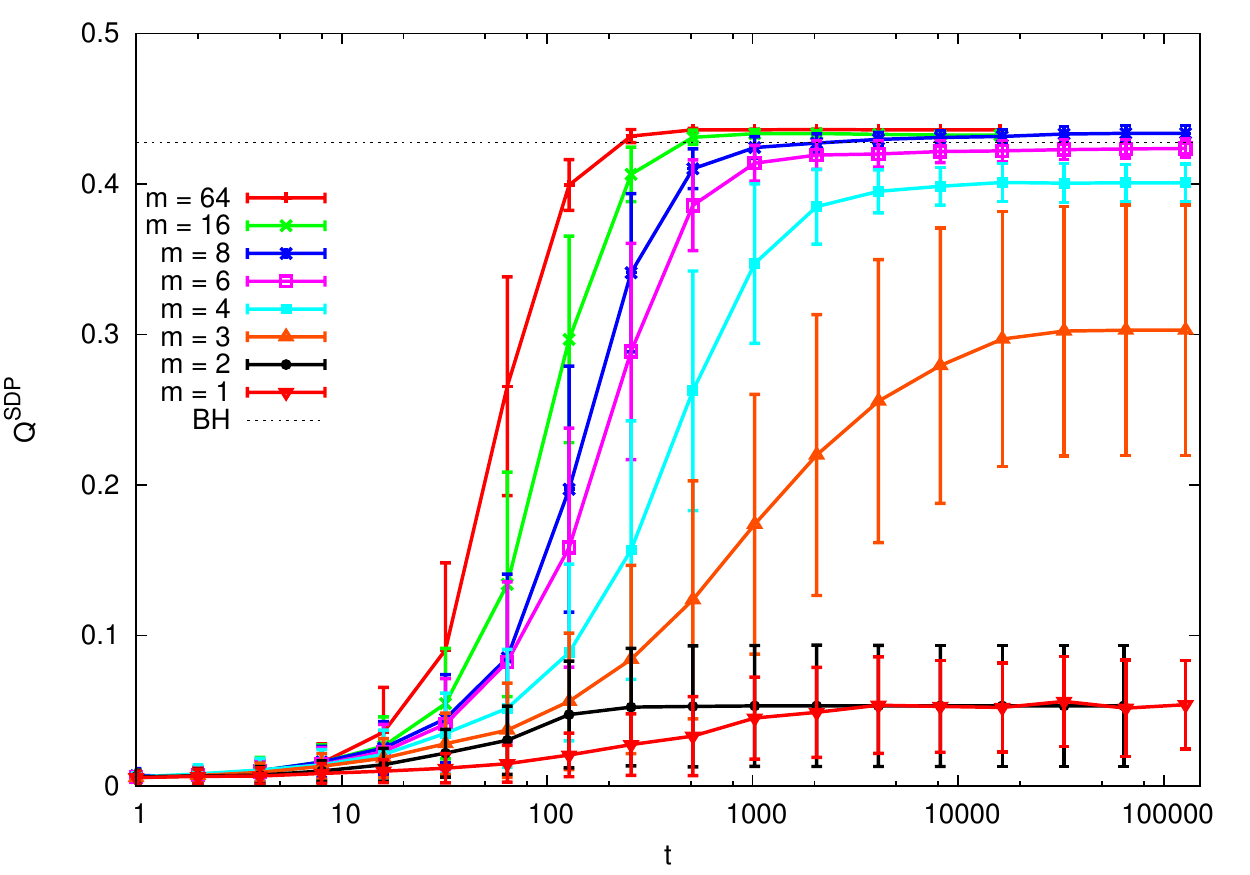}
\end{center}
\caption{\footnotesize Dependence on $m$ of the inferred overlap $\qsdp$ for $\lambda=1.1$. Error bars represent standard deviation in the population of 100 clones.
}
\label{fig:2}
\end{figure}

In Fig.~\ref{fig:2} we report the results for $\qsdp$ obtained on the same graph as in Fig.~\ref{fig:0} with many more values of $m$. Error bars represent the standard deviation over the 100 clones. We observe that, while for $m=1,2$ the algorithm is unable to detect any significant partition, already for $m=3$ the signal is substantial and eventually for $m \ge 8$ the algorithm achieves the optimal value for $\qsdp$. Being the graph size quite large ($30\,597$ vertices and $53\,161$ edges) is surprising that our algorithm works so nicely also for values of $m$ as small as $m=8$.

\begin{figure}[ht]
\begin{center}
\includegraphics[width=0.8\textwidth]{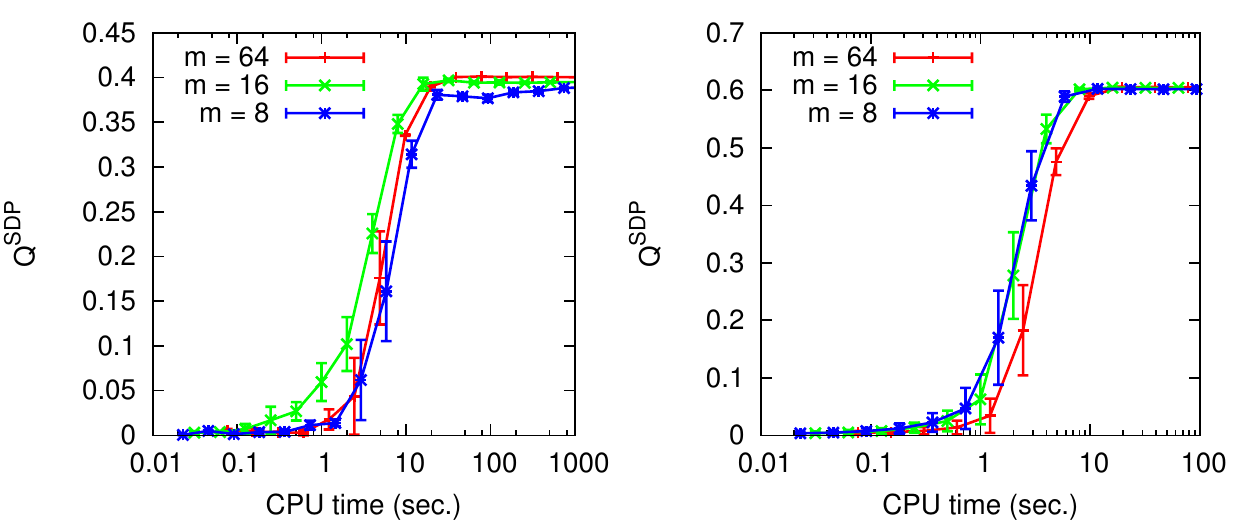}
\end{center}
\caption{\footnotesize Inferred overlap by our SDP-based greedy algorithm as a function of the real wall clock time. The two problems have been generated according to the SBM with $n=10^5$, $c=3$ and $\lambda=1.1$ (left) and $\lambda=1.2$ (right). Times have been measured in seconds on a personal laptop with a 2 GHz Intel Core i7 processor. Dependence on $m$ in running times is very weak.
}
\label{fig:CPUtime}
\end{figure}

Let us focus on values $m\ge 8$ and let us compute the real running time, that is the wall clock time our algorithm takes to maximize the objective function. This time is not exactly proportional to $m\,n$ because all the operations among the $m$-components vectors (e.g.\ sums and scalar products) can be highly optimized by modern compilers, and so we expect the real running time to grow sublinearly with $m$.
In Fig.~\ref{fig:CPUtime} we report the growth of $\qsdp$ for 2 graphs of initial size $n=10^5$ and $\lambda=1.1$ (left) and $\lambda=1.2$ (right); the sizes of their two cores are $77\,250$ nodes, $132\,578$ edges, and $77\,370$ nodes, $132\,890$ edges respectively. Error bars represent again the standard deviation over the 100 clones, while the CPU time has been measured for the simulation of a single clone on a personal laptop with a 2 GHz Intel Core i7 processor.
Looking at Fig.~\ref{fig:CPUtime} we notice that the differences in the real running time by varying the value of $m$ are not very significant, and so maybe working with a moderate value of $m$ is preferred, especially because the actual running time is very small: few seconds on a laptop to solve a problem with about $10^5$ variables!

\begin{figure}[ht]
\begin{center}
\includegraphics[width=0.8\textwidth]{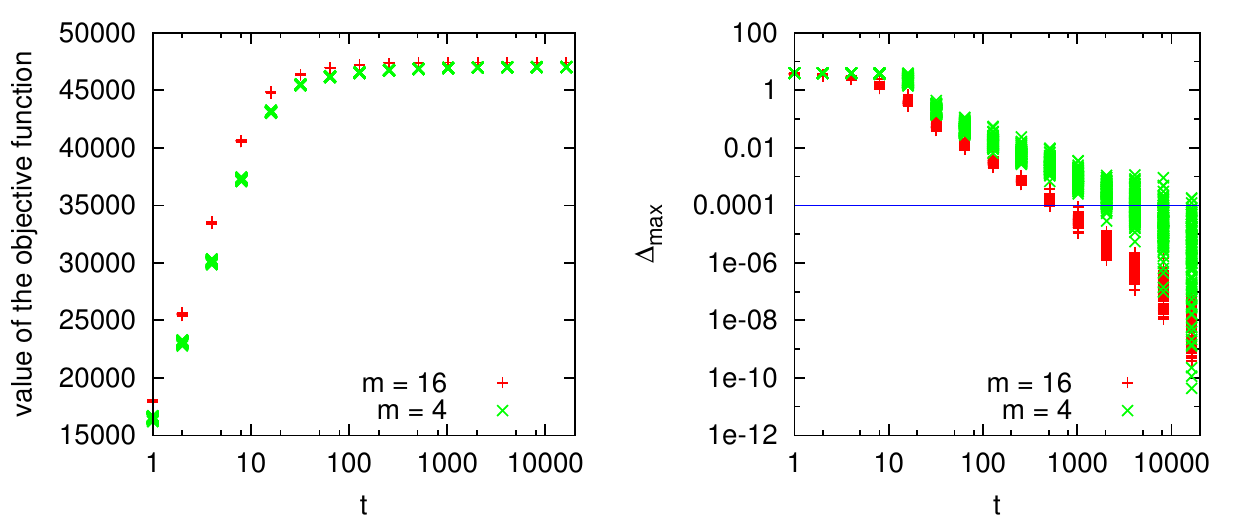}
\end{center}
\caption{\footnotesize The objective function and the maximum variation in a variable during the last step, $\Delta_\text{max}$, as a function of the number of algorithm steps. At each time we plot 100 points corresponding to different clones (although in the left panel they are hardly visible, because almost perfectly superimposed). The problem has size $n=40\,000$ and $\lambda=1.1$ (and is the same studied in Fig.~\ref{fig:histo}).
}
\label{fig:m4_m16}
\end{figure}

In Figs.~\ref{fig:0}, \ref{fig:2} and \ref{fig:CPUtime} we have always considered the overlap with the planted partition; however in practice such an overlap is unknown.
The only information available to us are the value of the objective function, $\Delta_\text{max}$ and the distances between clones (if we run more than one clone). The first two quantities are shown in Fig.~\ref{fig:m4_m16} for a graph generated with $n=40\,000$ and $\lambda=1.1$ (at each time we report the 100 measurements taken in different clones). From the previous analysis we known that $m=16$ may be a good value, while $m=4$ is definitely too small (see green and cyan lines in Fig.~\ref{fig:2}). However looking at the data in Fig.~\ref{fig:m4_m16} we do not see large differences in the objective function and only the fluctuations in $\tconv$, given by the time where $\Delta_\text{max}=10^{-4}$ (blue horizontal line in the right panel), are suggesting $m=4$ may be too small.

\begin{figure}[ht]
\begin{center}
$m = 16$

\includegraphics[width=0.82\textwidth]{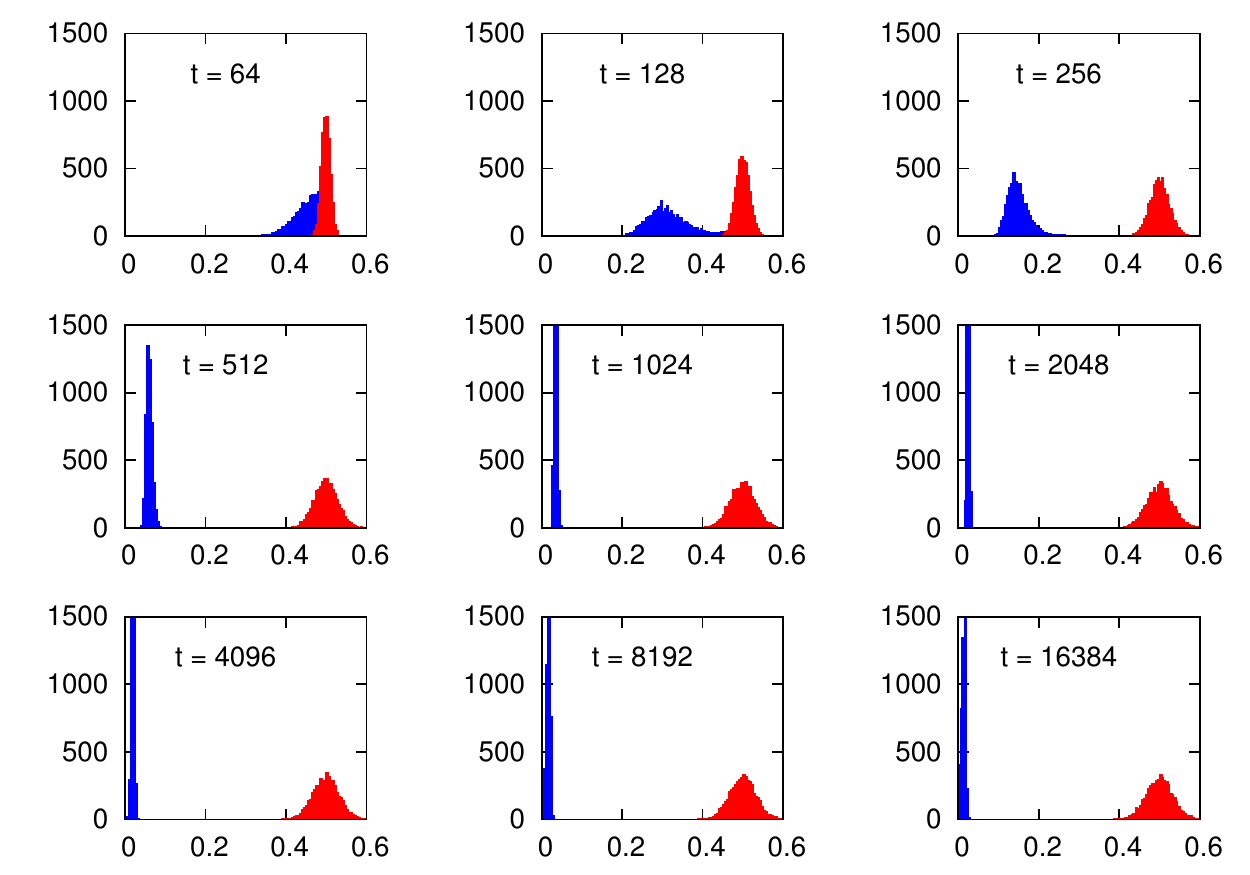}

\vspace{8mm}
$m = 4$

\includegraphics[width=0.82\textwidth]{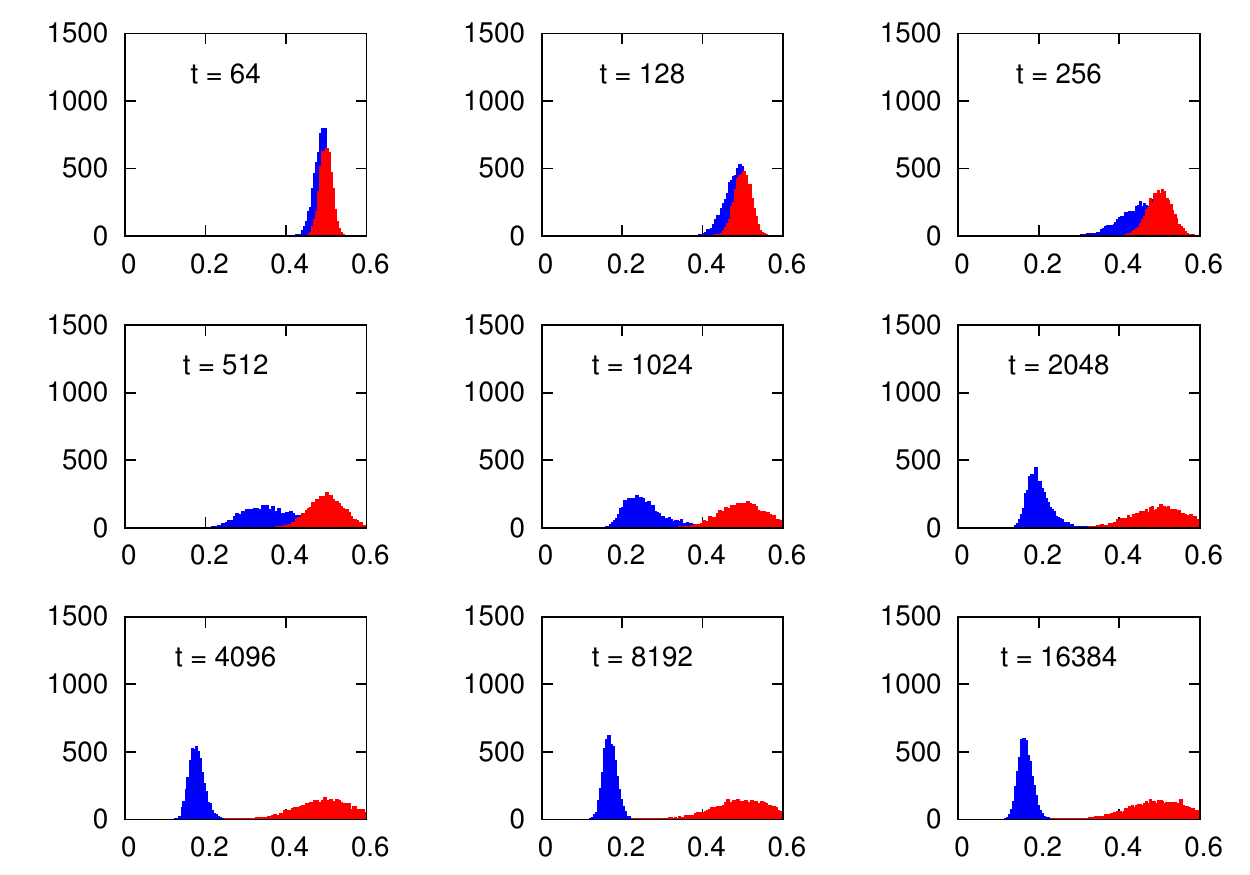}
\end{center}
\caption{\footnotesize Histograms of the 4950 distances among the 100 clones before (red) and after (blue) the eliminating the rotational symmetry. Each panel is for a different algorithm time; upper panels are for $m=16$ and lower ones are for $m=4$.
}
\label{fig:histo}
\end{figure}

Given that any clone eventually gets stuck in a maxima (either local or global), we would like to understand how likely the clones have reached the global maximum, without the need to run the algorithm with a larger value of $m$ (that would make convergence to the global maximum easier). To this end we measure the distances between any pair of clones, by measuring
\[
d = \frac12\left(1-\frac1n \sum_{i=1}^n \ux^{(1)}_i\cdot\ux^{(2)}_i\right)\,,
\]
where $\ubx^{(1)}$ and $\ubx^{(2)}$ are their configurations.
The idea is that if all the clones have reached the global maximum they are all very close by, while if the objective function is too rough and they get stuck in local maxima, then their distances remain strictly positive. There is one caveat in computing straightforwardly the distance between the clones: since the objective function is invariant under a rotation $R:\reals^m\to\reals^m$ applied to all the $n$ vectors and since the clones start from random configurations, we expect the clones to remain as orthogonal as possible, with a value of the distance $d \simeq 1/2$.
In Fig.~\ref{fig:histo} we plot in red the histograms of the 4950 distances between the 100 clones, measured during the running of the algorithm with $m=16$ (above) and $m=4$ (below). Indeed, we see that the peak of the distributions remain close to $d=1/2$ and provide no information at all about the real closeness of the clones.
In order to get a meaningful information we should break the rotational invariance, by rotating the configuration of each clone and minimize their distances. We have done this in the following way: for each clone, we have computed the empirical correlation function $\Sigma$ defined in \eqref{eq:Sigma}, the corresponding principal component $\uv_1$ and we have rotated the configuration such that $\uv_1 \parallel (1,0,\ldots,0)$ and all distances are smaller than $1/2$. The resulting distances are shown in Fig.~\ref{fig:histo} with the blue histograms: it is now clear that running the algorithm with $m=16$ makes all the clones converge to the same optimal configuration, while in the $m=4$ case the clones get stuck in local maxima and their distances converge to strictly positive values.

\section{Conclusions and perspectives}
\vspace{3mm}

We studied how the algorithm based on SDP for community detection introduced in Ref.~\cite{sdp} works on very large graphs generated according to the SBM.
We have shown the algorithm is very robust with respect to variations in the generative model: adding a good number of cliques only slightly degrades the algorithm performances (in contrast with spectral methods whose accuracy drops dramatically).
The present work demonstrates that surprisingly small values of $m$ may be sufficient for detecting the partition even in very large problems.
Running few clones (even just 2 clones) allows one to understand whether the global maximum of the objective function is achieved. Consequently one can choose whether to rerun the algorithm with a larger value for $m$.

There are many possible extensions of our work. We just mention the most straightforward ones. We have made some preliminary runs of the algorithm with $q=3$ and observed that 3 clusters of equal (or almost equal) size can be detected with a very good accuracy. However a systematic study of the performances of our algorithm for $q>2$ is still missing. Given that for $q>4$ the SBM undergoes a first order phase transition, it would be very interesting to study how the SDP relaxation in general and our algorithm in particular perform in detecting communities in that case. We think our algorithm can be also adapted to detect communities of different sizes by modifying the constraint of zero global magnetization; this modification is relevant to study real world problems.

\ack
\vspace{3mm}

A.J. and A.M. were partially supported by NSF grants CCF-1319979 and DMS-1106627 and the AFOSR grant FA9550-13-1-0036.

\section*{References}

\bibliographystyle{plain}
\bibliography{biblio}

\end{document}

%% file: commands.tex
\newcommand{\beq}{\begin{equation}}
\newcommand{\eeq}{\end{equation}}
\newcommand{\bea}{\begin{eqnarray}}
\newcommand{\eea}{\end{eqnarray}}
\newcommand{\<}{\langle}
\renewcommand{\>}{\rangle}
\newcommand{\E}{{\mathbb E}}

\newcommand{\cin}{c_{\rm in}}
\newcommand{\cout}{c_{\rm out}}
\newcommand{\tconv}{t_\text{conv}}

\def\bC{{\boldsymbol{C}}}

\def\reals{{\mathbb R}}

\def\bx{{\boldsymbol{x}}}

\def\hbx{\boldsymbol{\hat{x}}}

\def\qsdp{Q^{\mbox{\tiny{SDP}}}}
\def\lsdp{\lambda_{c}^{\mbox{\tiny{SDP}}}}

\def\xml{\boldsymbol{\hat{x}}^{\mbox{\tiny{ML}}}}
\def\xsdp{\boldsymbol{\hat{x}}^{\mbox{\tiny{SDP}}}}

\def\prob{{\mathbb P}}
\def\E{{\mathbb E}}

\def\<{\langle}
\def\>{\rangle}

\def\argmax{{\arg\!\max}}
\def\sign{{\rm sign}}

\def\uv{{\underline{v}}}
\def\ux{{\underline{x}}}
\def\ubx{{\underline{\boldsymbol{x}}}}


%% file: hd3-2015_v3.bbl
\begin{thebibliography}{10}

\bibitem{abbe2016exact}
Emmanuel Abbe, Afonso~S Bandeira, and Georgina Hall.
\newblock Exact recovery in the stochastic block model.
\newblock {\em Information Theory, IEEE Transactions on}, 62(1):471--487, 2016.

\bibitem{arora2005non}
Sanjeev Arora, Eli Berger, Elad Hazan, Guy Kindler, and Muli Safra.
\newblock On non-approximability for quadratic programs.
\newblock In {\em Foundations of Computer Science, 2005. FOCS 2005. 46th Annual
  IEEE Symposium on}, pages 206--215. IEEE, 2005.

\bibitem{boumal2014manopt}
Nicolas Boumal, Bamdev Mishra, P-A Absil, and Rodolphe Sepulchre.
\newblock Manopt, a matlab toolbox for optimization on manifolds.
\newblock {\em The Journal of Machine Learning Research}, 15(1):1455--1459,
  2014.

\bibitem{braun2006m}
A~Braun and T~Aspelmeier.
\newblock The m-component spin glass on a bethe lattice.
\newblock {\em Physical Review B}, 74(14):144205, 2006.

\bibitem{burer2003nonlinear}
Samuel Burer and Renato~DC Monteiro.
\newblock A nonlinear programming algorithm for solving semidefinite programs
  via low-rank factorization.
\newblock {\em Mathematical Programming}, 95(2):329--357, 2003.

\bibitem{clauset2004finding}
Aaron Clauset, Mark~EJ Newman, and Cristopher Moore.
\newblock Finding community structure in very large networks.
\newblock {\em Physical review E}, 70(6):066111, 2004.

\bibitem{decelle2011asymptotic}
Aurelien Decelle, Florent Krzakala, Cristopher Moore, and Lenka Zdeborov{\'a}.
\newblock Asymptotic analysis of the stochastic block model for modular
  networks and its algorithmic applications.
\newblock {\em Physical Review E}, 84(6):066106, 2011.

\bibitem{decelle2011inference}
Aurelien Decelle, Florent Krzakala, Cristopher Moore, and Lenka Zdeborov{\'a}.
\newblock Inference and phase transitions in the detection of modules in sparse
  networks.
\newblock {\em Physical Review Letters}, 107(6):065701, 2011.

\bibitem{fortunato2010community}
Santo Fortunato.
\newblock Community detection in graphs.
\newblock {\em Physics Reports}, 486(3):75--174, 2010.

\bibitem{goemans1995improved}
Michel~X Goemans and David~P Williamson.
\newblock Improved approximation algorithms for maximum cut and satisfiability
  problems using semidefinite programming.
\newblock {\em Journal of the ACM (JACM)}, 42(6):1115--1145, 1995.

\bibitem{guedon2015community}
Olivier Gu{\'e}don and Roman Vershynin.
\newblock Community detection in sparse networks via grothendieck’s
  inequality.
\newblock {\em Probability Theory and Related Fields}, pages 1--25, 2015.

\bibitem{holland1983stochastic}
Paul~W Holland, Kathryn~Blackmond Laskey, and Samuel Leinhardt.
\newblock Stochastic blockmodels: First steps.
\newblock {\em Social networks}, 5(2):109--137, 1983.

\bibitem{sdp}
Adel Javanmard, Andrea Montanari, and Federico Ricci-Tersenghi.
\newblock Phase transitions in semidefinite relaxations.
\newblock {\em Proceedings of the National Academy of Sciences},
  doi:10.1073/pnas.1523097113, 2016.

\bibitem{journee2010low}
Michel Journ{\'e}e, Francis Bach, P-A Absil, and Rodolphe Sepulchre.
\newblock Low-rank optimization on the cone of positive semidefinite matrices.
\newblock {\em SIAM Journal on Optimization}, 20(5):2327--2351, 2010.

\bibitem{kawamoto2015limitations}
Tatsuro Kawamoto and Yoshiyuki Kabashima.
\newblock Limitations in the spectral method for graph partitioning:
  Detectability threshold and localization of eigenvectors.
\newblock {\em Phys. Rev. E}, 91:062803, Jun 2015.

\bibitem{krzakala2013spectral}
Florent Krzakala, Cristopher Moore, Elchanan Mossel, Joe Neeman, Allan Sly,
  Lenka Zdeborov{\'a}, and Pan Zhang.
\newblock Spectral redemption in clustering sparse networks.
\newblock {\em Proceedings of the National Academy of Sciences},
  110(52):20935--20940, 2013.

\bibitem{lovasz1991cones}
L{\'a}szl{\'o} Lov{\'a}sz and Alexander Schrijver.
\newblock Cones of matrices and set-functions and 0-1 optimization.
\newblock {\em SIAM Journal on Optimization}, 1(2):166--190, 1991.

\bibitem{massoulie2014community}
Laurent Massouli{\'e}.
\newblock Community detection thresholds and the weak ramanujan property.
\newblock In {\em Proceedings of the 46th Annual ACM Symposium on Theory of
  Computing}, pages 694--703. ACM, 2014.

\bibitem{montanari2015semidefinite}
Andrea Montanari and Subhabrata Sen.
\newblock Semidefinite programs on sparse random graphs and their application
  to community detection.
\newblock ACM, 2016.

\bibitem{mossel2013proof}
Elchanan Mossel, Joe Neeman, and Allan Sly.
\newblock A proof of the block model threshold conjecture.
\newblock {\em arXiv preprint arXiv:1311.4115}, 2013.

\bibitem{newman2006modularity}
Mark~EJ Newman.
\newblock Modularity and community structure in networks.
\newblock {\em Proceedings of the National Academy of Sciences},
  103(23):8577--8582, 2006.

\bibitem{saade2014spectral}
Alaa Saade, Florent Krzakala, and Lenka Zdeborov{\'a}.
\newblock Spectral clustering of graphs with the bethe hessian.
\newblock In {\em Advances in Neural Information Processing Systems}, pages
  406--414, 2014.

\bibitem{schulke2015multiple}
Christophe Sch{\"u}lke and Federico Ricci-Tersenghi.
\newblock Multiple phases in modularity-based community detection.
\newblock {\em Physical Review E}, 92(4):042804, 2015.

\bibitem{zhang2014scalable}
Pan Zhang and Cristopher Moore.
\newblock Scalable detection of statistically significant communities and
  hierarchies, using message passing for modularity.
\newblock {\em Proceedings of the National Academy of Sciences},
  111(51):18144--18149, 2014.

\end{thebibliography}
